\documentclass{article}

\PassOptionsToPackage{numbers, compress}{natbib}

\usepackage[preprint]{neurips_2025}

\usepackage[utf8]{inputenc} 
\usepackage[T1]{fontenc}    
\usepackage{hyperref}       
\usepackage{url}            
\usepackage{booktabs}       
\usepackage{amsfonts}       
\usepackage{nicefrac}       
\usepackage{microtype}      
\usepackage{xcolor}         

\usepackage{wrapfig}
\usepackage{graphicx}
\usepackage{amsmath}
\usepackage{multirow}
\usepackage{adjustbox}
\usepackage{xcolor,colortbl}
\usepackage{tabularx}
\usepackage{authblk}
\usepackage{subcaption}

\usepackage{xcolor}
\definecolor{darkblue}{RGB}{25,25,180} 
\definecolor{darkred}{RGB}{180,0,0} 
\definecolor{darkgreen}{RGB}{0,180,0} 
\definecolor{midblue}{RGB}{25,25,112} 
\definecolor{darkyellow}{RGB}{240,180,0} 
\usepackage{hyperref}       
\hypersetup{
colorlinks=true,
linkcolor=darkred,
filecolor=darkred,      
citecolor=darkgreen,
}

\title{Not All Fakes are Equal: \\A Quality-Centric Framework for Deepfake Detection}

\author{
  \textbf{Wentang Song}\textsuperscript{1$^{*}$},
  \textbf{Zhiyuan Yan}\textsuperscript{2$^{*}$},
  \textbf{Yuzhen Lin}\textsuperscript{1},
  \textbf{Taiping Yao}\textsuperscript{2},
  \textbf{Changsheng Chen}\textsuperscript{1},
  \textbf{Shen Chen}\textsuperscript{2}, \\
  \textbf{Yandan Zhao}\textsuperscript{2},
  \textbf{Shouhong Ding}\textsuperscript{2},
  \textbf{Bin Li}\textsuperscript{1$^\dagger$}
}

\affil{
  {\tt 
  $*$ Equal Contributors, $\dagger$ Corresponding Authors
  }
  \par
  \textsuperscript{1} Shenzhen University,
  \textsuperscript{2} Tencent Youtu Lab
  \\
  {\tt 
  libin@szu.edu.cn
  }
}

\begin{document}
\definecolor{Gray}{gray}{0.9}

\maketitle

\begin{abstract}
Detecting AI-generated images, particularly deepfakes, has become increasingly crucial, with the primary challenge being the generalization to previously unseen manipulation methods.
This paper tackles this issue by leveraging the forgery quality of training data to improve the generalization performance of existing deepfake detectors.
Generally, the forgery quality of different deepfakes varies: some have easily recognizable forgery clues, while others are highly realistic. 
Existing works often train detectors on a mix of deepfakes with varying forgery qualities, potentially leading detectors to short-cut the easy-to-spot artifacts from low-quality forgery samples, thereby hurting generalization performance.
To tackle this issue, we propose a novel quality-centric framework for generic deepfake detection,  which is composed of a Quality Evaluator, a low-quality data enhancement module, and a learning pacing strategy that explicitly incorporates forgery quality into the training process.
Our framework is inspired by curriculum learning, which is designed to gradually enable the detector to learn more challenging deepfake samples, starting with easier samples and progressing to more realistic ones.
We employ both static and dynamic assessments to assess the forgery quality, combining their scores to produce a final rating for each training sample. 
The rating score guides the selection of deepfake samples for training, with higher-rated samples having a higher probability of being chosen.
Furthermore, we propose a novel frequency data augmentation method specifically designed for low-quality forgery samples, which helps to reduce obvious forgery traces and improve their overall realism.
Extensive experiments demonstrate that our proposed framework can be applied plug-and-play to existing detection models and significantly enhance their generalization performance in detection.
\end{abstract}

\section{Introduction}
\label{sec:Intro}
Image synthesis techniques have experienced rapid advancements in recent years, providing a new way of entertainment~\cite{gpt4o,gemini,yan2025gpt}. Unfortunately, this technology is also broadly misused to spread false misinformation and even bring political influence. To prevent the abuse of deepfake, it is increasingly crucial to develop a reliable deepfake detector.
Most earlier deepfake detectors~\cite{MesoNetCompactFacial2018afchar, FaceForensicsLearningDetect2019rossler, cao2021metric, gu2021spatiotemporal, khan2021video, kim2021cored} achieve satisfactory and promising detection results in the within-domain evaluation scenario, where the distributions of the training and testing data are similar. However, when encountering cross-domain evaluation, with previously unseen forgery methods or data sources, the performance of these models will drop significantly~\cite{FaceXRayMore2020li,luo2021generalizing,yan2023deepfakebench}. Thus, the generalization ability of these prior works is still limited and underexplored.



\begin{figure}[ht]
    \centering
    \begin{subfigure}[b]{0.65\textwidth}
        \includegraphics[width=\textwidth]{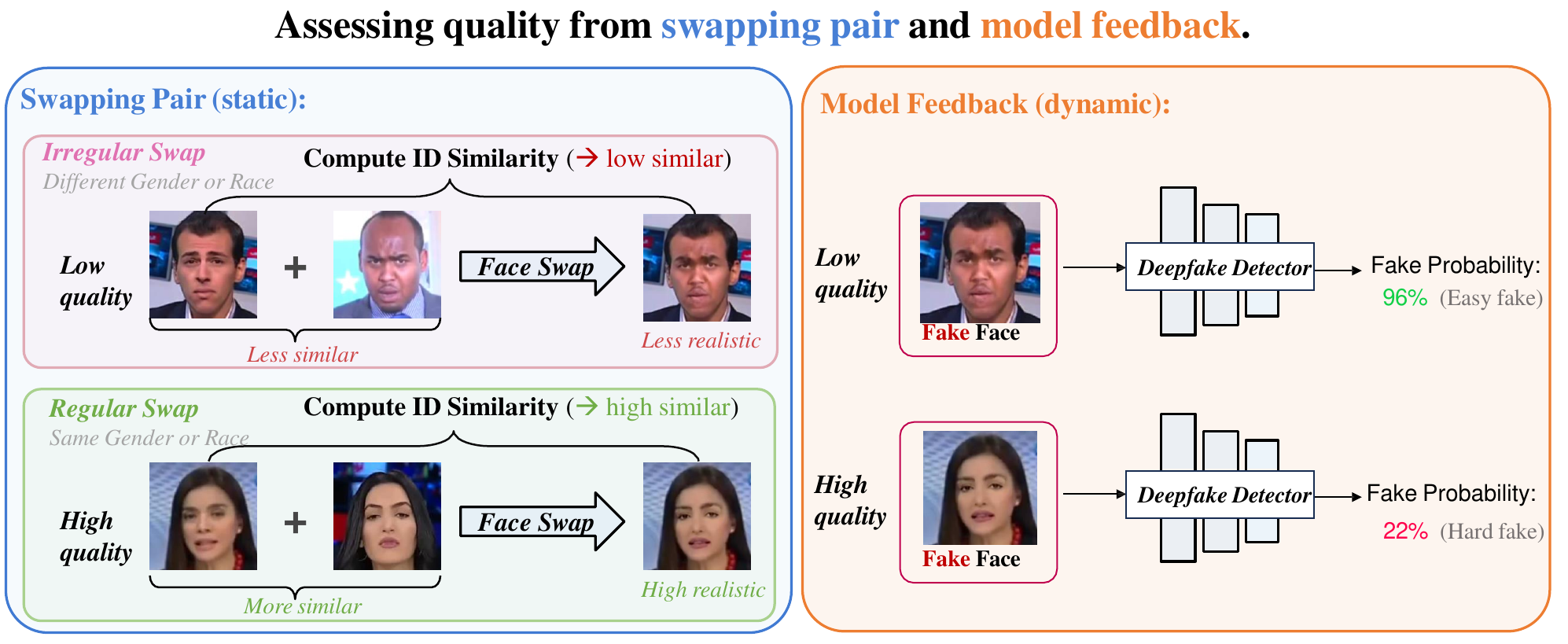} 
        \caption{Qualitatively assess the forgery quality from two views: swapping pairs (static) and model feedback (dynamic).}
        \label{fig:quality_concept}
    \end{subfigure}
    \hfill 
    \begin{subfigure}[b]{0.3\textwidth}
        \includegraphics[width=\textwidth]{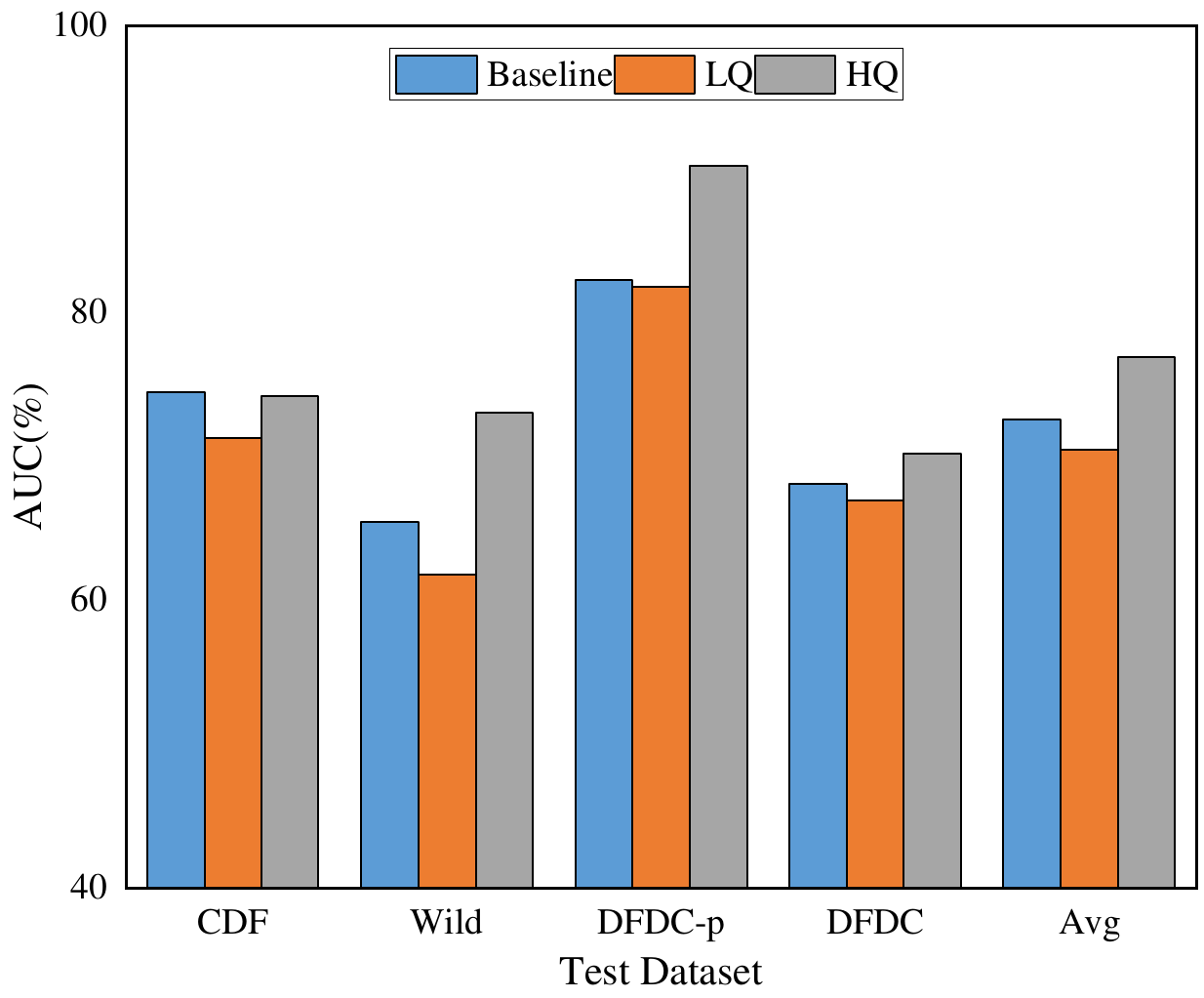} 
        \caption{Quantitative evaluation of model generalization performance across varying forgery quality levels. }
        \label{fig:arcface}
    \end{subfigure}
    \caption{Impact of forgery quality on deepfake detection. (a) Qualitative assessment of forgery quality from two perspectives: swapping pairs (static) and model feedback (dynamic); (b) Quantitative analysis of how different forgery quality levels affect model generalization performance.}
    \label{fig:quality}
\end{figure}

One reason for the generalization issue is the model's shortcuts~\cite{geirhos2020shortcut} on the easy-to-spot artifacts from the training samples with low forgery quality~\cite{yan2024transcending,DetectingDeepfakesSelfBlended2022shiohara,trinh2021examination}. For instance, the ``irregular swap" fake samples are frequently observed in FF++ (a common training set) but rarely in CDF-v2 (a common testing set). 
``Irregular swaps" was first noticed by \cite{trinh2021examination}, where face-swapping involves pairs with distinct facial attributes, such as differences in race or gender. 
These irregular swaps often severely distort the human facial structure, leading to lower-quality fake samples.
As illustrated in Figure~\ref{fig:quality_concept}, we see that when ``irregular swaps" occur (\textit{i.e.,} swapping a white face with a black face or a female face with a male face), the easy-to-recognize deepfake samples are generated.
To verify the impact of these lower forgery quality fakes on the model's generalization, we selectively choose samples with different forgery qualities for training the same detector (\textit{i.e.,} Xception~\cite{XceptionDeepLearning2017chollet}).

Results in Figure~\ref{fig:arcface} verify that these low static quality samples indeed hurt the model's generalization, especially in the Wild~\cite{WildDeepfakeChallengingRealWorld2020zi} and DFDC-P~\cite{DeepfakeDetectionChallenge2019dolhansky} datasets, where simply training with higher-quality samples can obtain about 10\% improvement in the generalization performance.

However, most existing methods often train detectors on a mix of qualities of deepfake data, which might inevitably struggle with this issue (similar to the ``baseline" in Figure~\ref{fig:arcface}), as they treat all data as the same and ignore the impact of different forgery qualities.


These observations motivate us to treat training data with forgery qualities differently. To this end, we need to address three critical considerations first:
\textbf{Question 1:} \textit{How can we assess the quality level of a sample?}
\textbf{Question 2:} \textit{How to deal with low-quality fake data?}
\textbf{Question 3:} \textit{How can we maximize the utilization of all types of data?}

For \textbf{Question 1}, we assess forgery quality from two perspectives—\textbf{static} (data-driven) and \textbf{dynamic} (model-driven)—as shown in Figure~\ref{fig:quality_concept}.
\textbf{From the static view}, we focus on swapping pairs in face-swapping. Deepfakes generated from similar identity pairs tend to be more realistic. High-quality samples show minimal visual artifacts and natural facial structure, while low-quality samples exhibit obvious distortions. To quantify this, we use a face recognition model (ArcFace~\cite{deng2019arcface}) to compute the cosine similarity between fake and corresponding real images (with aligned backgrounds); a lower similarity indicates lower static quality.
\textbf{From the dynamic view}, we evaluate how confidently the detection model predicts each fake during training. Lower prediction probabilities suggest harder-to-detect fakes, thus indicating higher dynamic quality (see Figure ~\ref{fig:quality_concept} for illustration).
By combining both static and dynamic scores, we define a Forgery Quality Score (FQS) to guide sample selection\footnote{Unless otherwise noted, low-quality and high-quality samples refer to those with low and high FQS, respectively.}.

For \textbf{Question 2}, instead of completely discarding low-quality deepfakes with obvious artifacts, we enhance their realism through a method called Frequency Data Augmentation (FreDA). FreDA improves low-quality samples by combining the low-frequency components from real faces (which preserve overall facial structure and semantic integrity) with the high-frequency components from fake faces (which contain more consistent and generalizable forgery cues). This fusion results in augmented samples with more natural facial structures and retained forgery signals, effectively transforming low-quality samples into more realistic, high-quality ones.
By doing so, FreDA reduces non-generalizable artifacts and strengthens the training data for better generalization.

For \textbf{Question 3}, to make full use of training samples with varying quality, we adopt a curriculum learning strategy that guides the model from easy to hard examples.
We first compute the Forgery Quality Score (FQS) for each sample and use it to control sampling during training—samples with higher FQS are more likely to enter the hard sample pool. 
As training progresses, we gradually reduce the number of training samples, shifting focus toward harder examples.
Meanwhile, we apply FreDA to enhance low-quality samples, increasing their difficulty and diversifying the training set.
This progressive training approach enables the model to benefit from the full spectrum of data quality.

Extensive experiments show it consistently improves both in-dataset and cross-dataset generalization across various deepfake detectors in a plug-and-play fashion.
Notably, our method outperforms the baseline~\cite{XceptionDeepLearning2017chollet, EfficientNetRethinkingModel2019tan, SwinTransformerV22022liu} by about 10\% on average across several widely-used evaluation datasets, such as Celeb-DF~\cite{CelebDFLargeScaleChallenging2020li} and DFDC~\cite{dolhansky2020deepfake}.

Overall, our main contributions are as follows:

\begin{itemize}

\item We design a \textbf{novel quality-centric training framework based on curriculum learning}, which is composed of a Quality Evaluator, a low-quality data enhancement module,  and a learning pacing strategy, encouraging the model to learn the deepfake artifacts gradually from easy to hard according to quality assessment. 

\item We propose a \textbf{quality assessment method combining both static and dynamic views} to obtain the Forgery Quality Score (FQS), which allows us to rank the hardness of the training samples, thereby implementing the training sample selection.

\item we propose \textbf{Frequency Data Augmentation (FreDA) to reduce the obvious forgery traces} and thereby enhance the realism of the augmented low-quality data, rather than directly discarding them. FreDA can reduce the traces of forgery in forged samples. Using samples processed by FreDA enhances the model's generalization.

\item Extensive experiments verify that the proposed framework \textbf{significantly improves the generalization} of the baseline model and can be applied in a \textbf{plug-and-play} manner.

\end{itemize}

\section{Related Works}
\label{Related_work}
Most detection works at an early stage, mainly focus on hand-crafted features, \textit{e.g.,} eye-blinking~\cite{IctuOculiExposing2018li}, inconsistencies~\cite{ExposingDeepFakes2019yang} of head poses~\cite{ExposingDeepFakeVideos2019li}, and other visual biological artifacts. 
With the rapid development of deep learning, data-driven-based detectors~\cite{FaceForensicsLearningDetect2019rossler} have shown better performance than the conventional hand-crafted approaches.  
However, these approaches often suffer from poor generalization when there is a distribution shift between training and testing forgeries~\cite{luo2021generalizing,yan2023deepfakebench,yan2024df40}.

Recent works have sought to improve generalization from different directions:

\textbf{1) Forgery Artifact Learning:} 
Several solutions, such as disentanglement learning~\cite{UCFUncoveringCommon2023yan, ExploringDisentangledContent2022liang}, reconstruction learning~\cite{EndtoEndReconstructionClassificationLearning2022cao, RepresentativeForgeryMining2021wang} and inconsistent learning~\cite{UIAViTUnsupervisedInconsistencyAware2022zhuang, NoiseBasedDeepfake2023wang, tian2024learning, yin2023dynamic, TALLThumbnailLayout2023xu, ImplicitIdentityLeakage2023dong}, have been proposed to learn general forgery artifacts. These methods aim to isolate the common characteristics of manipulated content, irrespective of the specific forgery technique employed. 
To illustrate, UCF~\cite{UCFUncoveringCommon2023yan} proposed a disentanglement-based framework to eliminate the overfitting of both forgery-irrelevant content and forgery-specific artifacts.
CADDM~\cite{ImplicitIdentityLeakage2023dong} used the identity inconsistency between the inner and outer faces for detection.
Additionally, 
However, these approaches mostly operate at the spatial level but overlook frequency-level artifacts, which have been demonstrated to be crucial for detection~\cite{qian2020thinking}.
To this end, several studies have incorporated frequency information to enhance the performance of detectors~\cite{durall2020watch, qian2020thinking, liu2021spatial, DynamicGraphLearning2023wang, AdaptiveFaceForgery2022song}. 
Additionally, several recent works~\cite{kong2022detect, 10138555, 10023530, yan2024effort} have proposed to leverage both local and global interactions to more accurately capture forgery traces. 
Furthermore, \cite{DualContrastiveLearning2022sun, luo2023beyond, SeeABLESoftDiscrepancies2023larue} leveraged contrastive learning to improve the detection performance of deepfake detectors.

\textbf{2) Forgery Augmentation:} Forgery augmentation techniques have proven to be highly effective in enhancing the generalization ability of deepfake detectors by encouraging the model to learn generic forgery artifacts. Pioneering works such as BI (Face X-Ray)~\cite{FaceXRayMore2020li} synthesized blended faces by replicating blending artifacts between pairs of pristine images with similar facial landmarks. I2G~\cite{zhao2021learning} employed a similar strategy, leveraging pair-wise self-consistency learning to identify inconsistencies within generated fake images. Later efforts, like SBI~\cite{DetectingDeepfakesSelfBlended2022shiohara}, focused on self-blending, selecting the same source and target faces to train models, while ProDet~\cite{cheng2024can} explored the way to fuse augmented forgery data with actual deepfake data more effectively.
More recently, \cite{AltFreezingMoreGeneral2023wang} and \cite{yan2024generalizing} have explored video-level blending for data augmentation, achieving generalizable results in detecting video deepfakes.

\textbf{3) Quality-aware deepfake detection:} 
Another notable direction is to train the detector with different image qualities (combining both low-quality compression images and high-quality non-compression ones), aiming to learn the quality-irrelevant forgery artifacts~\cite{le2023quality}.
Recent works~\cite{kim2024correlation, song2024towards} proposed to investigate the correlation between the detection performance and the image quality.
Additionally, \cite{lee2022bznet} proposed a super-resolution model for improving low-quality deepfake detection.
However, most existing works define the ``quality" as the \textbf{image quality}, \textit{e.g.,} whether the image is blurry or not.
Although our framework is also quality-based, our focus is more on the \textbf{forgery quality}, \textit{i.e.}, whether the deepfake image is realistic or not, rather than the image quality.
In our framework, we propose to assess the forgery quality from two distinct aspects: swapping pairs and model feedback. We then design a curriculum-learning-based method to leverage samples with varying forgery qualities dynamically for training, guiding the detector to learn the samples from easy to hard gradually.

\section{Method}
\label{Method}

In this section, we introduce the proposed quality-centric framework. Our framework is inspired by curriculum learning that explicitly encourages the detection model to learn forgery artifacts gradually from easy to hard based on FQS. 
It emphasizes three key components: firstly, the evaluation of data quality; secondly, the data augmentation on low-quality data; and thirdly, the progressive use of quality-based data during training.
To achieve this, we carefully design three specified modules to implement the framework:
\textbf{1)} FQS is proposed as a metric to assess sample hardness during model training. It consists of two main components: \textit{static scores and dynamic scores}. Static score (related to data) provides initial guidance at the start of model training. The weight of the static score gradually decreases as the training progresses. Dynamic score (related to the model) is continuously updated based on the model's feedback, to better screen samples throughout the training process;
\textbf{2)} A FreDA module is proposed to enhance the forgery quality (``realism") of the augmented low-quality fake samples;
\textbf{3)} A pacing function is presented to control the pace of presenting data from easy to hard according to FQS. Each module will be detailed subsequently.
\begin{figure*}[t]
	\centering
	\includegraphics[width=\linewidth]{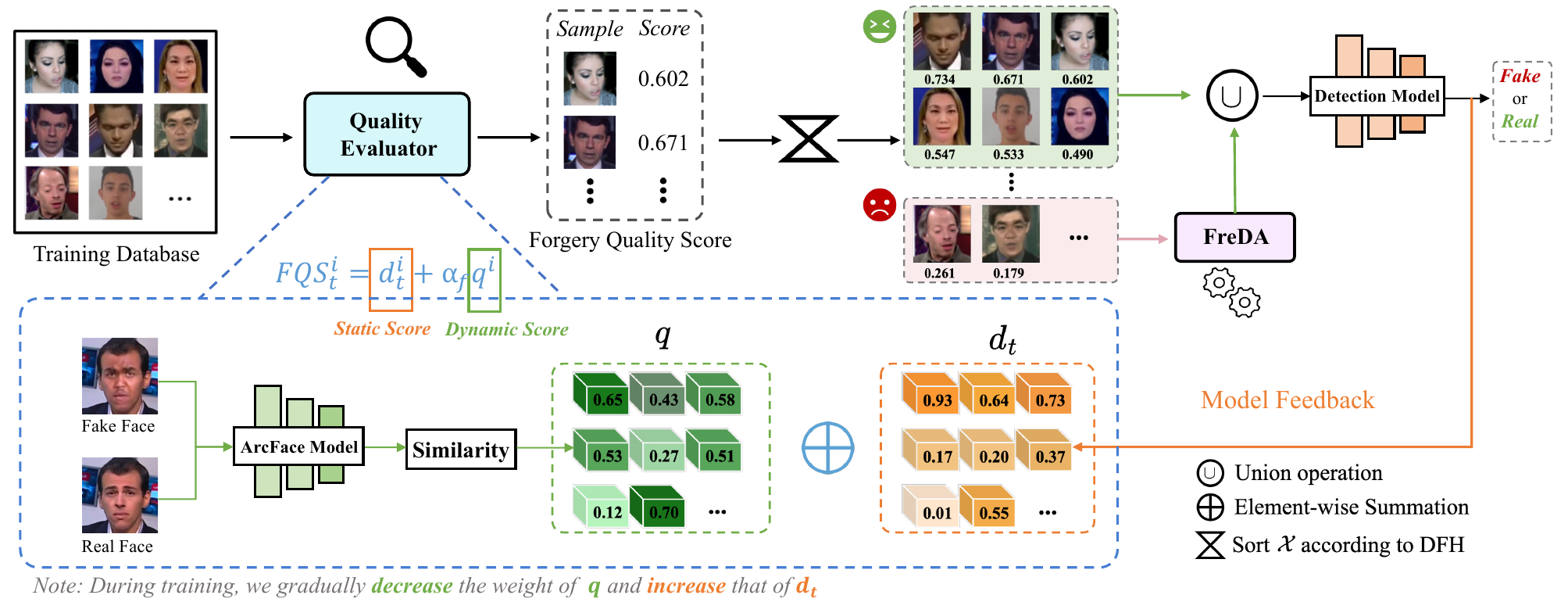}
	\caption{Overall pipeline of the proposed method. To improve the existing data, we classify the samples through the Quality Evaluator module. For low-quality data, we use the FreDA module as shown Figure~\ref{freda}, to improve the quality of those samples.}
    \vspace{-1em}
	\label{overview}
\end{figure*}

\subsection{Forgery Quality Score (FQS)}
The hardness score plays a key role in curriculum learning as it describes the relative ``hardness" of each sample in deepfake detection. In this work, we propose the FQS, which considers the dynamic model behavior and the static facial data quality.

\noindent \textbf{Static quality.} 
The face recognition network learns rich facial feature representations through a large amount of external data. 
The features extracted through the face recognition network can capture detailed information and specific face attributes in the face image. 
High-quality deepfake images may have more accurate facial alignment, more realistic texture details, and more natural expression changes. 
Their features are more similar to the original images, causing the features extracted by the face network to be more similar to the original image features. 
On the contrary, low static quality deepfake images may have distortion, blur, or obvious forged details, resulting in features extracted by the face network being less similar to the original image features. 
 For commonly used training datasets, there is a correspondence between deepfake videos and real videos.
Taking advantage of such a fact, we input the fake image and its corresponding real image into a pre-trained face recognition network (Arcface), and calculate the cosine similarity of the features obtained by the model. 
The higher the similarity, the better the quality of the sample, and vice versa. 

For an input fake image $x_{f}^{i}$, we obtain its corresponding real image $x_{r}^{i}$, input it into the face recognition network G($\cdot $), and obtain two features $g_{f}^{i}$, $g_{r}^{i}$:$g_{f}^{i}=G(x_{f}^{i})$, $g_{r}^{i}=G(x_{r}^{i})$.
After that, we calculate the cosine similarity between the two features as the ID similarity score $q^{i}$ of $x_{f}^{i}$:
$q^{i}=Cosine(g_{f}^{i},g_{r}^{i})$.
The similarity score $q^i$ is used as the static quality for the forgery quality evaluation, as the more similar face-swapping pairs, the more realistic the deepfake is created.

\noindent \textbf{Dynamic quality.} 
Let $ \mathcal{X} = {(x^{i},y^{i} )}_{i=1}^{N} $ be the training dataset with $N$ samples, $x^{i}$ and $y^{i}$ represent the $i$-th data and its ground-truth label, respectively. Let $f(\cdot,\theta_{t})$ be the deepfake detector model with the parameter $\theta_{t}$ at $t$-th epoch. 
We regard the loss (\textit{i.e.}, the binary cross-entropy loss denoted as $l(f(x^{i}, \theta_{t-1}), y^{i})$) of $i$-th sample at $t-1$-th epoch as an indicator for the current hardness of this sample judged by the current state of the model before conducting the training step.  Thus, we propose the instantaneous hardness $s_{t}^{i}$ that normalized the current loss with the learning rate $\eta_{t}$, formulated as:
	$s_{t}^{i}=l(f(x^{i};\theta_{t-1} ),y^{i})\cdot \eta_{max}/\eta_{t},$
where $\eta_{max}$ is the max learning rate during the training. 
We measure the dynamic hardness ${d}_{t}^{i}$ through a moving average of the instantaneous hardness ${s}_{t}^i$ over training history, defined and computed recursively as:
\begin{equation}\label{eq2}
	{d}_{t}^i=
	\begin{cases}
		\gamma \times s_{t}^i + (1-\gamma) \times {d}_{t-1}^i,& \text{if} \ \, i\in H_t\\
		{d}_{t-1}^i,& \text{otherwise},
	\end{cases}
\end{equation}
where $\gamma \in [0,1]$ is a discount factor, and $H_t$ is the subsets of hard samples at $t$-th epoch selected by the pacing function. 
Finally, we perform a weighted summation of the static quality and dynamic quality to obtain the FQS, formulated as:
\begin{equation}\label{eq1}
	FQS_{t}^i={d}_{t}^i+\alpha_{f}q^i,
\end{equation}
where $\alpha_{f}$ is a balance weight. As the training progresses, $\alpha_{f}$ will gradually decrease.

\subsection{Frequency Data Augmentation (FreDA)}
\begin{wrapfigure}{r}{0.4\textwidth}
    \vspace{-2em}  
    \includegraphics[width=0.4\textwidth]{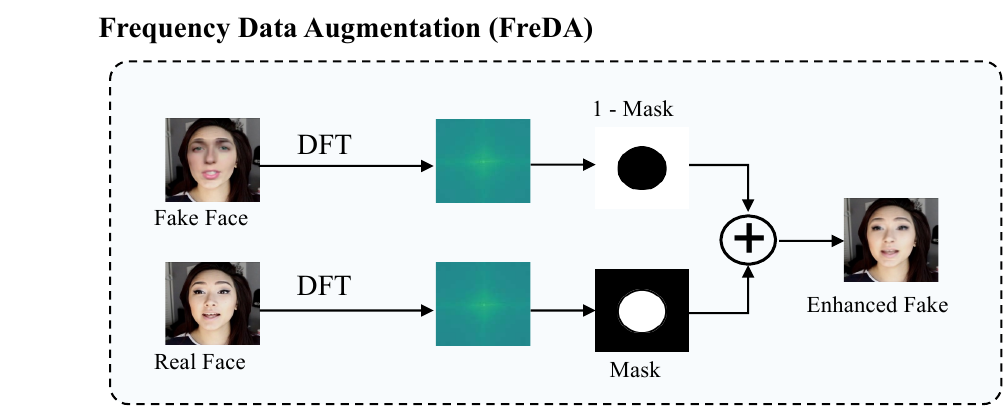}
    \vspace{-1em}  
    \caption{For low-quality data, we employ the FreDA module to augment their forgery quality by reducing the easily recognizable artifacts and enhancing their realism.
    }
    \label{freda}
\end{wrapfigure}

For a face image, its low-frequency information typically corresponds to the overall structure, outline, and general range of the face, while the high-frequency information represents the texture and subtle features of the face. In the case of fake samples, discrepancies with real samples are predominantly concentrated in high-frequency areas, indicating that tampering traces are often found in high-frequency components. 

For these low-quality samples, the facial structure is severely damaged. We can use the low-frequency part of the corresponding real sample of the fake sample as the low-frequency component of the blended image, and use the high-frequency part of the fake sample as the high-frequency component of the blended image. In this way, the obtained sample will preserve the facial structure without destruction, while also retaining traces of tampering-related forgery.

Specifically, given a sample $x\in\mathbb{R}^{H\times W\times C}$ we perform a 2D fast Fourier transform (FFT) for each channel independently to get the corresponding frequency representation $F$.
Further, we denote with $M$ a mask, whose value is zero except for the center region:
\begin{equation}
    \begin{array}{l} 
M(u,v)=
    \begin{cases}
1, (u,v)\in  [c_h-r:c_h+r,c_w-r:c_w+r]\\ 
  0, others \\ 
\end{cases}  , 
\end{array} 
\end{equation}
where  $(c_h,c_w)$ is the center of the image and $r$ indicates the frequency threshold that distinguishes between high- and low-frequencies of the original image. Given a low-quality sample $x_{f}$, we first find its corresponding real sample $x_{r}$, and then perform FFT on these two images to obtain $F_{f}$ and $F_{r}$. We splice the high-frequency information of the fake sample $x_{f}$ and the low-frequency information of the real sample $x_{r}$ to obtain $F_a$. The formula is expressed as follows:
\begin{equation}
    F_a=F_{r}\otimes M+F_{f}\otimes (1-M),
\end{equation}
where $\otimes$ denotes element-wise multiplication. 
Perform iFFT changes on the obtained $F_a$ and convert it to the RGB domain, and then obtain the FreDA image $x_a$.
The frequency representation $F$ can be transferred to the original RGB space via an inverse FFT (iFFT).
Compared with $x_{f}$, $x_a$ has a relatively complete facial structure while retaining the traces of forgery. 

The proposed FreDA focuses on frequency-level augmentation and distinguishes itself from previous studies in the following ways.
\textbf{Compared to frequency-based detectors}:
FreDA aims to reduce the obvious forgery artifacts of a less realistic deepfake image and transform it into a realistic one at the frequency level, while previous frequency detectors~\cite{qian2020thinking,liu2021spatial} mainly mine the frequency signals (\textit{e.g,} high-frequency components) as the auxiliary information of the original RGB and encourage the detector to learn spatial-frequency anomalies.
\textbf{Compared to augmentation-based detectors}: 
FreDA is targeted on deepfake data and designed to reduce the obvious fake artifacts of the deepfake data at the frequency level, while previous augmentation-based detectors~\cite{FaceXRayMore2020li,zhao2021learning,SelfSupervisedLearningAdversarial2022chen,DetectingDeepfakesSelfBlended2022shiohara} mostly target the real data and simulate the fake blending artifacts to transform the real data into the (more realistic) pseudo-fake data.
Overall, our proposed FreDA is a novel frequency-level augmentation method that is distinct from both existing frequency detectors and augmentation-based detectors.

\subsection{Learning Pacing}
To control the learning pace of presenting data from easy to hard, we design a pacing function to determine the sample pool $\mathcal{X}^{\prime}_{t}$ of training data according to FQS. 
Like human education, if a teacher presents materials from easy to hard in a very short period, students will become confused and will not learn effectively.
Thus, we define a pacing sequence $\mathbf{T}=[T_{0}, \dots , T_{N}]$ to represent $n+1$ milestones (\text{i.e.}, $n$ episodes) in total training epoch $T$.
During the first $T_{0}$ epochs, we utilize all the samples in $\mathcal{X}$ for the warm-up training. After epoch $T_{0}$, we only change the size of $\mathcal{X}^{\prime}_{t}$ at every milestone $T_{n}$. Specifically, at each epoch $t$ of episode $n$, we select $k_{n}$ samples with top FQS values in $\mathcal{X}$ (\text{i.e.}, hardest samples) as a hard sample pool $H_{t}$. Along with the training, we reduce the size of $H_{t}$ by $k_{n}\gets \alpha _{\beta}\times k_{n-1}$ with discount factor $\alpha_{\beta}$ to make it gradually focus on harder samples.
To further enlarge the diversity of the data, we also select the $E$ samples with bottom FQS values in $\mathcal{X}$ (\text{i.e.}, easiest samples) as an easy sample pool $E_{t}$ and then conduct $FreDA$ on them.
Then, we get the sample pool $\mathcal{X}^{\prime}_{t}$ by mixing the $H_{t}$ and $FreDA(E_{t})$, \text{i.e.}, $\mathcal{X}^{\prime}_{t}\gets H_{t}\cup FreDA(E_{t})$. Finally, we sample a mini-batch in $\mathcal{X}^{\prime}_t$ and then conduct a vanilla training step to update the model parameter as $\theta_{t}$.


\section{EXPERIMENTS}
\label{others}
\subsection{Experiment Settings}
\noindent \textbf{Datasets.} In this paper, we train all models on the \textbf{FaceForensics++ (FF++)}~\cite{FaceForensicsLearningDetect2019rossler} dataset. 
To demonstrate the performance of our proposed method in cross-dataset settings, four
additional datasets are adopted, i.e.,
Celeb-DFv2 \textbf{(CDF)}~\cite{CelebDFLargeScaleChallenging2020li},
Deepfake Detection Challenge Preview \textbf{(DFDC-p)}~\cite{DeepfakeDetectionChallenge2019dolhansky},
Deepfake Detection Challenge Public Test Set \textbf{(DFDC)}~\cite{dolhansky2020deepfake}, and  
WildDeepfake \textbf{(Wild)}~\cite{WildDeepfakeChallengingRealWorld2020zi}.
For cross-manipulation evaluation, we utilize two Deepfake datasets: FF++ (train a model on one method from FF++ and test it across all four methods (DF, F2F, FS, and NT)) and \textbf{DF40} (train a model on  FF++ and test it across the methods in DF40) .

\noindent \textbf{Implementation details.} As for pre-processing, we utilize MTCNN to detect and crop the face regions (enlarged by a factor of 1.3) from each video frame and resize them to 256 $\times$ 256. For each video, we select every 10 frames to form the training dataset. We adopt Swin-Transformer-V2 Base (Swin-V2)~\cite{SwinTransformerV22022liu} as the default backbone network, and the parameters are initialized by the weights pre-trained on the ImageNet1k. 
We employed the SGD optimizer with a cosine learning rate scheduler with $\eta_{max}=0.1$ (used in \ref{eq1}). 
We set 20 epochs for totally training, the pacing sequence $\mathbf{T}=[2,5,8,12,15],$ and hyper-parameters $\gamma=0.9$ (used in \ref{eq2}), $\alpha_{\beta}=0.9$, $E=1000$.
The parameter $\alpha_{f}$ is initialized at 0.5 and undergoes a reduction by half at each milestone.

\noindent \textbf{Evaluation Metrics.}
In this work, we mainly report the area under the ROC curve (AUC) to compare with prior works.
The video-level results are obtained by averaging predictions over each frame on an evaluated video. 

\subsection{Performance Comparisons}

\subsubsection{Cross-Datasets Comparisons}
The cross-dataset evaluation is still a challenging task because the unknown domain gap between the training and testing datasets can be caused by different source data, forgery methods, and/or post-processing. In this part, we evaluate generalization performance in a cross-dataset setting. Specifically, our models were trained on the FF++ and tested on  CelebDF, Wild, DFDC-p, and DFDC, respectively.
We compare our proposed method against twenty-three state-of-the-art DeepFake detectors in the past three years:  RECCE~\cite{EndtoEndReconstructionClassificationLearning2022cao}, SLADD~\cite{SelfSupervisedLearningAdversarial2022chen},
UIA-ViT~\cite{UIAViTUnsupervisedInconsistencyAware2022zhuang}, 
PEL~\cite{gu2022exploiting},
UCF~\cite{UCFUncoveringCommon2023yan}, SFDG~\cite{DynamicGraphLearning2023wang}, NoiseDF~\cite{NoiseBasedDeepfake2023wang}, FoCus~\cite{tian2024learning}, LSDA~\cite{yan2024transcending}, 
ProgressDet~\cite{cheng2024can},
DCL~\cite{DualContrastiveLearning2022sun},  
CD-NeT~\cite{AdaptiveFaceForgery2022song},  
EFNB4 + SBIs~\cite{DetectingDeepfakesSelfBlended2022shiohara}, 
CFM~\cite{luo2023beyond},  
SeeABLE~\cite{SeeABLESoftDiscrepancies2023larue},  TALL~\cite{TALLThumbnailLayout2023xu}, AUNet~\cite{AUNetLearningRelations2023bai}, 
PIM~\cite{guan2024improving}, 
DFFC~\cite{song2024towards} and
FreqBlender~\cite{zhou2024freqblender}. 
The experimental results in terms of frame-level and video-level AUC are shown in Table \ref{tab:cross-dataset}.
\begin{wrapfigure}{r}{0.5\textwidth}
    \begin{subfigure}[b]{0.24\textwidth}
        \includegraphics[width=\textwidth]{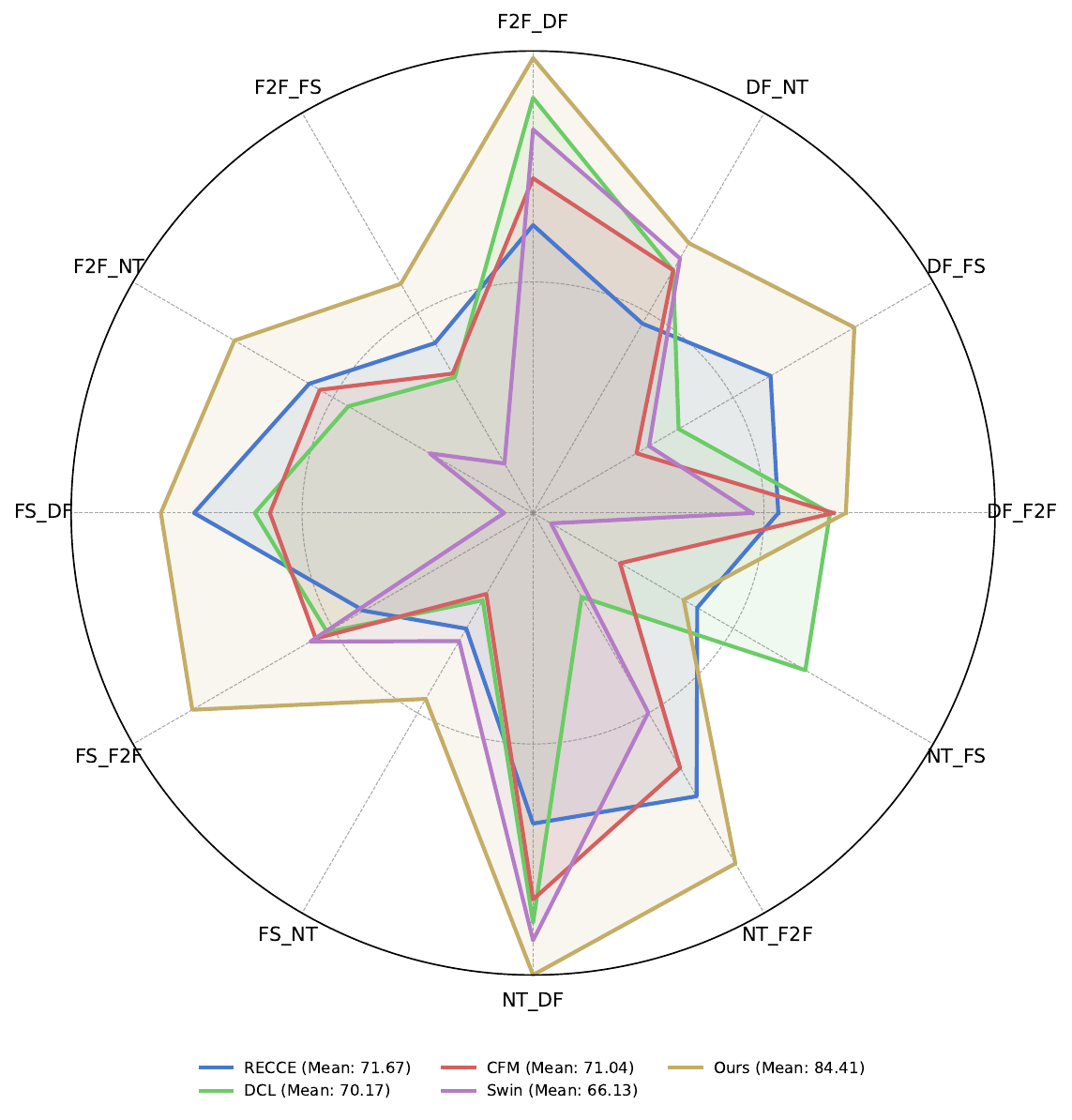}
        \caption{The performance of cross-manipulation methods on the FF++.}
        \label{fig:CM_FF}
    \end{subfigure}
    \hfill
    \begin{subfigure}[b]{0.24\textwidth}
        \includegraphics[width=\textwidth]{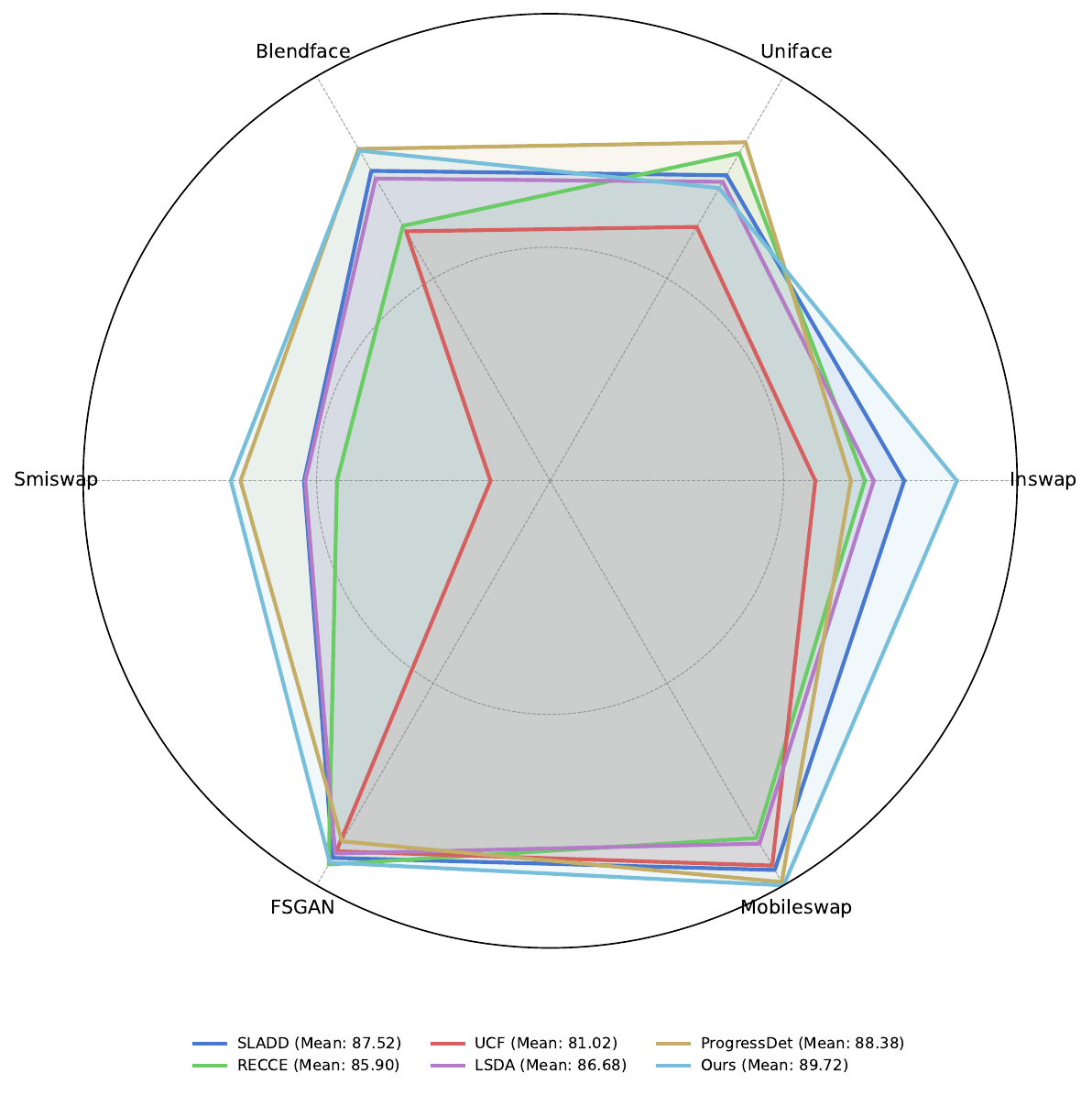}
        \caption{The performance of cross-manipulation methods on the DF40.}
        \label{fig:CM_DF40}
    \end{subfigure}
    \caption{Performance comparison of cross-manipulation with other SOTA detectors under the latest DF40 dataset.}
    \label{fig:cross-manipulation}
\end{wrapfigure}
It is evident that the proposed method lead to superior performance compared to other models in most cases, achieving the overall best results. All models are trained on the FF++ (C23) dataset for a fair comparison. 
We observe that our method surpasses the leading competitors in both frame-level and video-level evaluations. 
For instance, our approach surpasses
TALL~\cite{TALLThumbnailLayout2023xu}, which also employs Swin Transformer as its backbone network, by around 4\% and 6\% when testing on CDF and DFDC. The experimental results clearly highlight the significance of our method in enhancing the generalization capabilities of deepfake detectors.

\begin{table*}[tb!]
\centering
\caption{Frame-level (left) and Video-level (right) AUC (\%) of cross-datasets performances compared with SOTA methods.  The top two results are highlighted, with the best in \textbf{bold} and the second-best \underline{underlined}. `*' indicates our reproductions.}
\label{tab:Protocol-1}
\scalebox{0.65}{ 
\small 
\setlength{\tabcolsep}{2pt} 
\renewcommand{\arraystretch}{1.0} 
\begin{tabular}{c  c | c c c c c | c  c  c |c c c c c} 
\toprule
\multicolumn{7}{c}{\textbf{Frame-Level AUC}} & & \multicolumn{7}{c}{\textbf{Video-Level AUC}} \\ 
\cmidrule(lr){1-7} \cmidrule(lr){9-15}
\textbf{Method} & \textbf{Venues} & \textbf{CDF} & \textbf{Wild} & \textbf{DFDC-p} & \textbf{DFDC} & \textbf{Avg} & &
\textbf{Method} & \textbf{Venues} & \textbf{CDF} & \textbf{Wild} & \textbf{DFDC-p} & \textbf{DFDC} & \textbf{Avg} \\ 
\midrule
\midrule
RECCE        & CVPR 2022 & 82.30 & 75.63 & 74.91 & 69.06 & 75.48 & & DCL        & AAAI 2022 & 82.30 & 71.14 & - & 76.71 & 76.72 \\
SLADD             & CVPR 2022 & 83.70 & 69.02 & 75.60 & \underline{77.20} & 76.39 & & CD-Net        & ECCV 2022 & 88.50 & -- & -- & 77.00 & -- \\
UIA-ViT    & ECCV 2022 & 82.41 & -- & 75.80 &-- & -- & & EFNB4 + SBIs   & CVPR 2022 & 93.18 & -- & 86.15 & 72.42 & 83.92 \\
PEL    & AAAI 2022 & 69.18 & 67.39 & -- & 63.31 & 66.63 & & CFM   & TIFS 2023 & 89.62 & \underline{82.27} & -- & \underline{80.22} & 84.04 \\
UCF      & ICCV 2023 & 75.27 & 77.40 & 75.94 & 71.91 & 75.13 & & SeeABLE           & CVPR 2023 & 87.30 & -- & 86.30 & 75.90 & 83.17 \\
SFDG          & CVPR 2023 & 75.83 & 69.28 & -- & 73.64 & 72.92 & & TALL   & ICCV 2023 & 90.79 & -- & -- & 76.78 & -- \\
NoiseDF            & AAAI 2023  & 75.89 & -- & -- & 63.89 & -- & & AUNet           & CVPR 2023 & 92.77 & -- & 86.16 & 73.82 & 84.25 \\
FoCus             & TIFS 2024 & 76.13 & \underline{73.31} & 76.62 & 68.42 & 73.62 & & PIM           & TIFS 2024 & 87.34 & 72.28 & 80.69 & 68.04 & 77.09 \\
LSDA   & CVPR 2024 & 83.00 & -- & \underline{81.50} & 73.60 & \underline{79.37} & & DFFC        & ICASSP 2024 & 82.26 & 71.75 & \underline{90.63}  & 72.37 & 79.25 \\
ProgressDet     & NeurIPS 2024 & \underline{84.48} & -- & 81.16 & 72.40 & 79.35 & & Freq         & NeurIPS 2024 & \underline{94.59} & -- & 87.56 & 74.59 & \underline{85.58} \\
\midrule
\midrule
\multirow{2}{*}{\textbf{Ours}} & \multirow{2}{*}{--} & 
\textbf{89.18} & \textbf{80.74} & \textbf{90.94} & \textbf{77.65} & \textbf{84.63} & &
\multirow{2}{*}{\textbf{Ours}} & \multirow{2}{*}{--} & 
\textbf{94.75} & \textbf{83.77} & \textbf{93.41} & \textbf{82.04} &\textbf{88.49} \\
& & \textcolor[rgb]{0.10,0.52,0.27}{(+4.70\%)} & \textcolor[rgb]{0.10,0.52,0.27}{(+7.43\%)} & \textcolor[rgb]{0.10,0.52,0.27}{(+9.44\%)} & \textcolor[rgb]{0.10,0.52,0.27}{(+0.45\%)} & \textcolor[rgb]{0.10,0.52,0.27}{(+5.26\%)} & & & & \textcolor[rgb]{0.10,0.52,0.27}{(+0.16\%)} & \textcolor[rgb]{0.10,0.52,0.27}{(+1.50\%)} & \textcolor[rgb]{0.10,0.52,0.27}{(+2.78\%)} & \textcolor[rgb]{0.10,0.52,0.27}{(+1.82\%)} & \textcolor[rgb]{0.10,0.52,0.27}{(+2.91\%)} \\
\bottomrule
\end{tabular}
} 
\label{tab:cross-dataset}
\end{table*}

\subsubsection{Cross-manipulation Comparisons}

In real-world scenarios, predicting the methods used for face manipulation is often challenging. Therefore, it is crucial for our model to exhibit strong generalization capabilities to unseen manipulations. 
We conducted experiments using the FF++(HQ) dataset. We trained a model on one method from FF++ and tested it across all four methods (DF, F2F, FS, and NT). We select three state-of-the-art (SOTA) methods for comparison, including RECCE~\cite{EndtoEndReconstructionClassificationLearning2022cao}, 
DCL~\cite{DualContrastiveLearning2022sun},  
CFM~\cite{luo2023beyond}, and our baseline model Swin-V2~\cite{SwinTransformerV22022liu}. 
As shown in Fig~\ref{fig:CM_FF}, our proposed method can improve the average AUC by more than 10\% compared with the baseline Swin-V2. 

In addition, we also trained on FF++, and tested on six datasets in DF40, namely Inswap, Uniface, Blendface, Smiswap, FSGAN, and Mobileswap, the results are shown in Fig~\ref{fig:CM_DF40}. The comparative methods include advanced ones such as SLADD~\cite{SelfSupervisedLearningAdversarial2022chen},
RECCE~\cite{EndtoEndReconstructionClassificationLearning2022cao},
UCF~\cite{UCFUncoveringCommon2023yan},
LSDA~\cite{yan2024transcending} and
ProgressDet~\cite{cheng2024can}.
Our method has improvements over the SOTA method, which shows that our method has good generalization when detecting unseen manipulation methods.

\subsection{Ablation Study}

In this section, we conduct ablation studies on different components of the proposed model. All ablation studies are carried out on the HQ of FF++.
Note that we aim to verify the effectiveness of the proposed methods targeted on the three research questions.
We validate the effectiveness of both the static and dynamic forgery quality assessment strategies for \textbf{Question 1}, as given in Section \ref{score_ab}.
We validate the effectiveness of the proposed FreDA for \textbf{Question 2}, as given in  Section \ref{freda_ab}.
We validate the effectiveness of the proposed curriculum learning framework for \textbf{Question 3}, as given in Section \ref{other_strategy}.

\subsubsection{Scalability and extensibility of FreDA} 
\label{freda_ab}
\begin{figure}[ht]
    \centering
    \begin{subfigure}[b]{0.45\textwidth}
        \includegraphics[width=\textwidth]{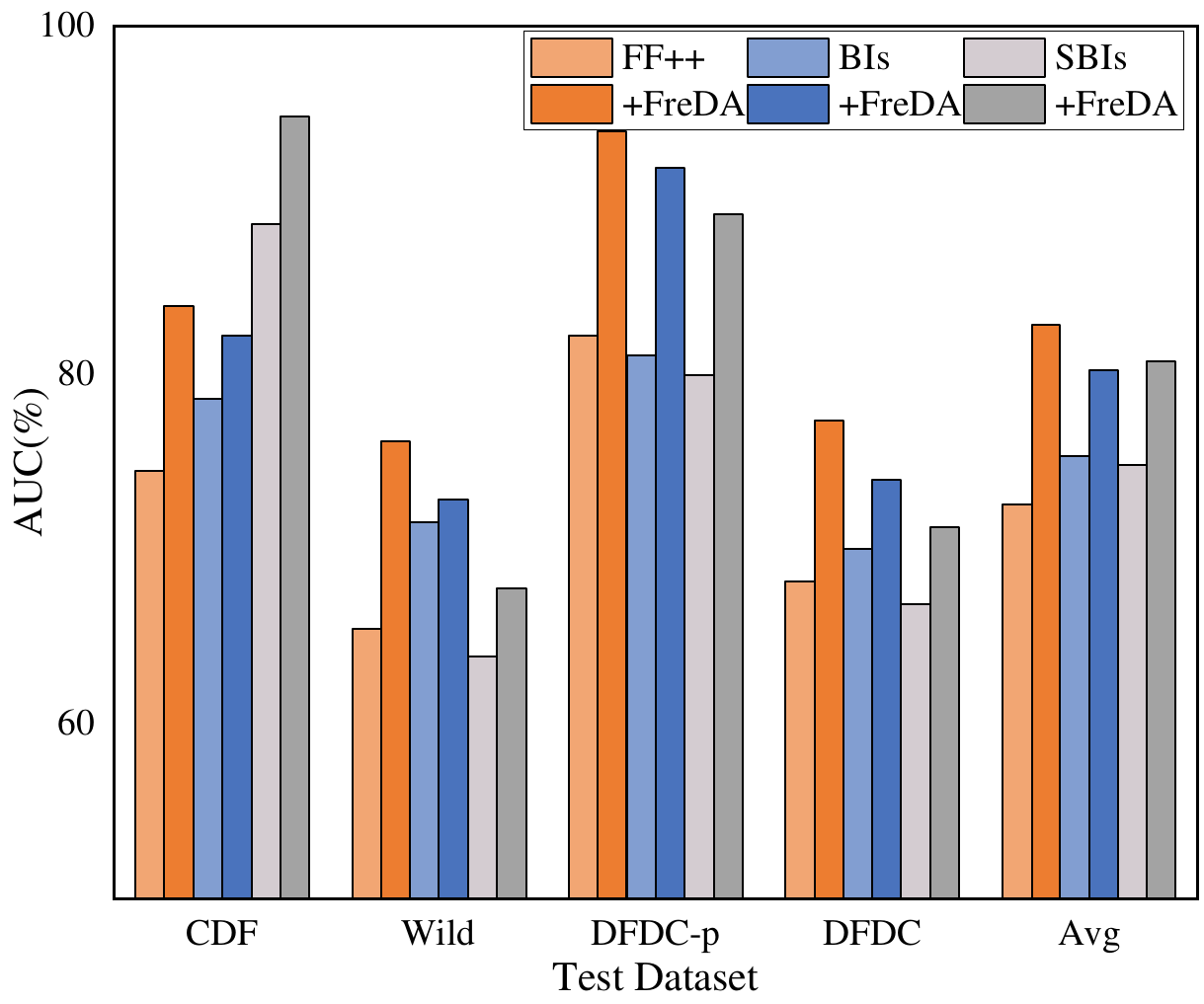} 
        \caption{FreDA can improve the generalization of models trained on all types of data.}
        \label{fig:FreDA_Scalability}
    \end{subfigure}
    \hfill 
    \begin{subfigure}[b]{0.45\textwidth}
        \includegraphics[width=\textwidth]{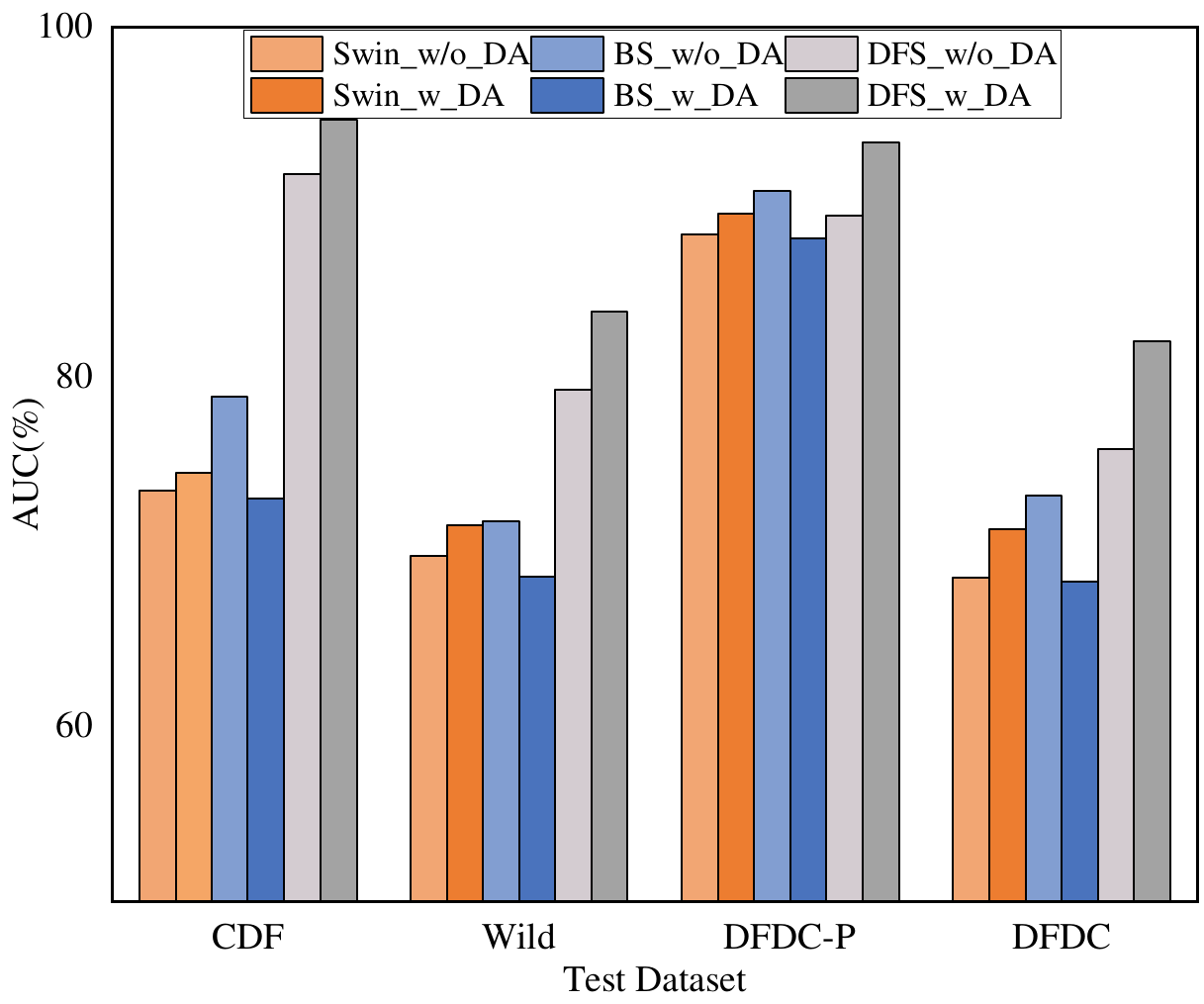} 
        \caption{Video-level AUC(\%) on cross-datasets performance of different training strategies.}
        \label{fig:different_strategy}
    \end{subfigure}
    \caption{Ablation studies on the proposed FreDA and Pacing Function. (a) Exploration of FreDA's extensibility; (b) The impact of different training strategies under the cross-dataset setting.}
    \label{fig:ablation_fig}
\end{figure}
FreDA augments data through splicing at the frequency domain level, and it can improve any data. Here, we combine FreDA with FF++, Blended images (BIs), and Self-Blended images (SBIs) to improve the data of BIs and SBIs and use these data for model training. The results are shown in Fig \ref{fig:FreDA_Scalability}. For the results of BIs and SBIs, we use the code provided by the author of the paper to reproduce. Since their paper uses RAW data for training, this paper uses The HQ version of FF++, so the performance is better than that of the original paper. 
The results show that the performance of combining BIs and FreDA is 4\% higher in terms of AUC than that of BIs on average, the performance of combining SBIs and FreDA is 5\% higher than that of SBIs on average, and the performance of combining FF++ and FreDA is 10\% higher than that of FF++ on average. And the detection performance of all datasets has been improved.
In addition, it is worth noting that BIs and SBIs have fewer forgery traces than FF++, while FreDA has a greater improvement effect on FF++, which also verifies that FreDA has a greater effect on improving samples with obvious forgery traces.
This also confirms the rationality of using FreDA on simple samples.

\subsubsection{Comparison with other training strategies}
\label{other_strategy}
In this part, we compared our method with other training strategies, including the vanilla training and BabyStep~\cite{CurriculumLearning2009bengio, cirik2016visualizing} which is the simplest CL strategy that utilizes a static pre-defined hardness. We conducted BabyStep by introducing the static  score as the pre-defined hardness and utilizing the pacing setting in~\cite{PowerCurriculumLearning2019hacohen}. 
Furthermore, we also investigated the impacts of data augmentation. As shown in Figure~\ref{fig:different_strategy}, we observe that both the CL paradigm and data augmentation can improve performance compared to vanilla training in most cases. However, introducing data augmentation in BabyStep suffers from performance degradation, as it makes the augmented data fail to match its pre-defined static hardness measure. This demonstrates that the dynamic CL strategy with data augmentation (\textit{i.e.}, our method) is beneficial for generic deepfake detection.
The key finding here is that the combination of a dynamic CL approach and data augmentation, as used in our method, is more effective than a static CL strategy with data augmentation. This suggests that the flexibility and responsiveness of the dynamic CL approach are crucial for leveraging data augmentation to improve detection performance.


\subsubsection{The impact of different definitions of hardness scores}
\label{score_ab}

The definition of hardness score plays a crucial role in curriculum learning, so in this section, we use different kinds of hardness scores to conduct ablation experiments. 
We show the performance of vanilla training (baseline), using only Dynamic Quality (DQ), and using DQ in combination with Static Quality (SQ) as the hardness definition. The results are shown in Table~\ref{tab:scores}. 
It can be seen that by using DQ as the definition of hardness scores, the generalization ability of the model can be improved, achieving a 5.17\% gain in terms of AUC than the model obtained by vanilla training.
After combining DQ and SQ, the performance of the model is further improved, with an average detection performance of 88.49\%.
It shows that introducing the prior knowledge of the face recognition model (Arcface) helps to improve the definition of hardness score. Arcface score can improve the hardness score from different perspectives, which also confirms our motivation to introduce static scores.

\begin{table}[t]
    \centering
    \caption{Ablation studies of the proposed modules. (a) Ablations on different hardness scores and (b) ablations of different data augmentations for easy samples with lower hardness scores.}
    
    \begin{tabular}{@{}cc@{}}
    \begin{minipage}{0.48\textwidth}
        \centering
        \subcaption{Effect of hardness scores.}
        \begin{adjustbox}{width=\textwidth}
        \begin{tabular}{ccccccc}
        \toprule
        \multicolumn{2}{c}{Score} & \multicolumn{4}{c}{Test Dataset} \\
        \midrule
         SQ    & DQ    & CDF   & Wild  & DFDC-p & DFDC  & Avg \\
        \midrule
               --   &    --   & 74.53  & 71.56  & 89.37  & 71.30  & 76.69  \\
               --  &   \checkmark   & 89.86  & 72.19  & 89.85  & 75.55  & 81.86  \\

            \checkmark      & \checkmark      & \textbf{94.75} & \textbf{83.77} & \textbf{93.41} & \textbf{82.04} & \textbf{88.49} \\
        \bottomrule
        \end{tabular}%
        \label{tab:scores}
        \end{adjustbox}
        
    \end{minipage} &
    \begin{minipage}{0.48\textwidth}
        \centering
        \subcaption{Effect of operations on easy samples.}
        \begin{adjustbox}{width=\textwidth}
        \begin{tabular}{cccccccc}
        \toprule
        \multicolumn{3}{c}{DA} & \multicolumn{5}{c}{Test Dataset} \\
        \midrule
        w/o DA & Or DA  & FreDA & CDF   & Wild  & DFDC-p & DFDC  & Avg \\
        \midrule
        \checkmark     &    --   &  --     & 91.63  & 70.41  & 86.37  & 74.16  & 80.64  \\
           --   & \checkmark      &     --  & 89.54  & 79.26  & 89.26  & 75.89  & 83.49  \\
           --   &     --  & \checkmark      & \textbf{94.75} & \textbf{83.77} & \textbf{93.41} & \textbf{82.04} & \textbf{88.49} \\
        \bottomrule
        \end{tabular}%
        \label{tab:DA}
        \end{adjustbox}
    \end{minipage} \\
    \end{tabular}
\end{table}

\subsubsection{The impact of the operation on easy samples}
\label{easy_sample_ab}
In this section, we evaluate the impact of different data augmentations on the model's generalization. We use three settings: no data augmentation (w/o DA), ordinary DA (such as JPEG compression, Gaussian Blur, Affine, or DA), and FreDA. 
The results are shown in Table~\ref{tab:DA}.
It can be seen that the generalization ability of the model can be improved by using simple data augmentation, with an average improvement of 2.85\%, but there is an obvious decrease in CDF, with a performance reduction of 2.09\%. 
When FreDA is performed on simple samples, the performance is improved across the board, with an average improvement of 7.85\% compared to the setting without data augmentation.
The results prove that data augmentation has a positive impact on the generalization ability of the model, and also proves the effectiveness of the FreDA we proposed.

\section{Conclusion}
This paper proposes a quality-centric framework that explicitly utilizes the forgery quality for generic deepfake detection.
The Forgery Quality Score (FQS) is assessed statically via swapping pairs and dynamically based on the model's feedback.
Inspired by curriculum learning, we perform sample selection via FQS during training: a higher quality sample has a higher probability of being sampled for training the model.
This manner aims to encourage the model to gradually learn more challenging samples, avoiding overfitting to easy-to-recognizable samples, thereby improving the generalization performance.
Furthermore, for these low-quality fake data, we propose a frequency data augmentation method called FreDA, which aims to reduce the easy artifacts of these samples and enhance their realism at the frequency level.
Extensive experiments prove that our method can be applied to other methods in a plug-and-play manner, and can largely improve the generalization of the baseline model.


\bibliographystyle{plain}
\bibliography{ref}

\end{document}